\documentclass[10pt,twocolumn,letterpaper]{article}

\usepackage[pagenumbers]{cvpr} 

\usepackage{graphicx}
\usepackage{amsmath}
\usepackage{amsthm}
\usepackage{amssymb}
\usepackage{booktabs}
\usepackage{multirow}
\usepackage{algorithm,algorithmic}
\usepackage[table,xcdraw]{xcolor}
\usepackage{enumitem}

\usepackage{graphicx}
\usepackage{subcaption}
\usepackage[most]{tcolorbox}
\usepackage{makecell} 
\usepackage{subcaption} 
\newtheorem{definition}{Definition}[section]

\providecommand{\keywords}[1]{%
  \small	
  \textbf{\textit{Keywords---}} #1
}

\usepackage[pagebackref,breaklinks,colorlinks]{hyperref}

\usepackage[capitalize]{cleveref}
\crefname{section}{Sec.}{Secs.}
\Crefname{section}{Section}{Sections}
\Crefname{table}{Table}{Tables}
\crefname{table}{Tab.}{Tabs.}

\def\cvprPaperID{*****} 
\def\confName{CVPR}
\def\confYear{2025}

\usepackage[T1]{fontenc}
\begin{document}

\title{
  A Modern Look at Simplicity Bias in Image Classification Tasks
}

\author{Xiaoguang Chang \\ 
School of Cyber Science and Engineering, Southeast University\\
Nanjing, China \\
{\tt\small xg\_chang@seu.edu.cn}
\and
Teng Wang\\
School of Automation, Southeast University\\
Nanjing, China \\
{\tt\small wangteng@seu.edu.cn}
\and
Changyin Sun\\
School of Automation, Southeast University\\
Nanjing, China \\
{\tt\small cysun@seu.edu.cn}
}
\maketitle
\begin{abstract}

The simplicity Bias (SB) of neural networks, i.e.\ their tendency to represent simple functions, is a key factor in their generalization capabilities. Recent studies show that an excessive SB may harm performance on complex tasks, and the need for this bias varies across tasks. Many of these studies focus on simple models or synthetic tasks. It remains challenging to measure the SB in large models and little is known about the relevance of the SB to various image classification tasks.
In this paper, we investigate the relationship between the SB in CLIP models and their performance across image classification tasks. First, we theoretically analyze the potential limitation of existing measures of complexity that have been used to characterize small models. To address this, we propose a frequency-aware measure capturing finer-grained SB differences. We validate this measure on CLIP models subjected to two recent SB-modulation methods, demonstrating that it is more informative and consistent than previous measures. Second, we examine the relation between the SB of those models and their performance across a range of image classification tasks, including zero-shot and fine-tuning settings. These experiments reveal a range of behaviors. For example, a stronger SB correlates with a better performance on OOD generalization than on adversarial robustness. These results highlight the benefits of aligning a model's inductive biases with the characteristics of the target task.

\end{abstract}

\section{Introduction}




Neural networks (NNs) exhibit a phenomenon known as \emph{simplicity bias} (SB)~\cite{valle2019deep,huh2021low,shah2020pitfalls}, referring to their tendency to
fit their training data with ``simple'' functions.
This bias stems from architectural choices and the choice of activation functions in particular~\cite{Teney_2024_CVPR},
while the optimization by gradient descent can further reinforce the effect~\cite{kalimeris2019sgd}. The SB plays a crucial role in the ability of NNs to achieve good generalization~\cite{park2022vision,arpit2017closer,valle2019deep,kalimeris2019sgd,Teney_2024_CVPR}.

\begin{figure}[t]
    \centering

    \includegraphics[width=1\linewidth]{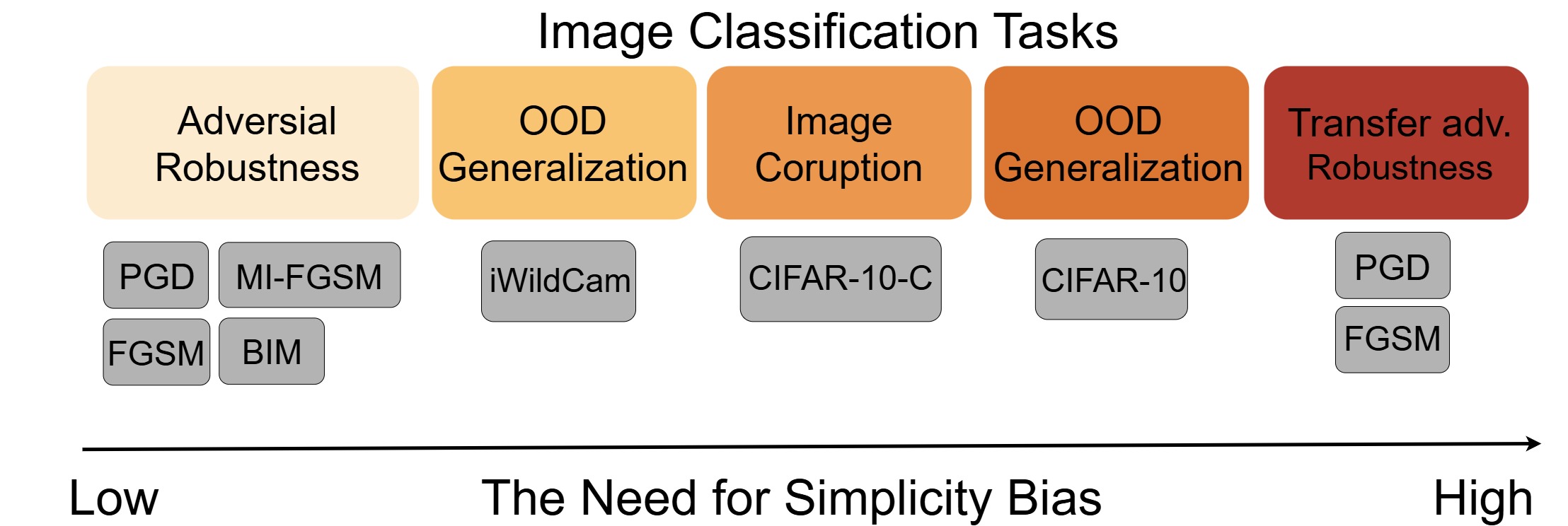}
    \caption{Comparison of fine-tuning tasks (with corresponding benchmarks or evaluation methods shown in gray). From left to right, tasks exhibit increasing preference for models with stronger SB.
}
        \label{fig:abstractfig}
\end{figure}

While the SB is generally beneficial for generalization, it may also lead to shortcut learning and ultimately harm model performance~\cite{Teney_2024_CVPR,Teney_2025_CVPR,shah2020pitfalls,vasudeva2023mitigating}. For instance, Shah et al.~\cite{shah2020pitfalls} use synthetic datasets to demonstrate that an excessive simplicity bias contributes to poor OOD generalization and adversarial vulnerability. Other recent studies aim to better understand the impacts of simplicity bias across various scenarios. Teney et al.~\cite{Teney_2024_CVPR} discover that simplicity bias can be modulated by activation functions and LayerNorm weights, and investigate the effects of varying simplicity bias on highly controllable datasets.
In follow-up work~\cite{Teney_2025_CVPR}, the authors modulate the SB by learning new activation functions,
and show that various tasks benefit from a SB of various strength.
This conclusion aligns with the widely accepted view that a model's inductive bias should match the characteristics of the target task.



Most studies on the SB focus on small models or on controllable tasks. Hence its role in modern larger-scale settings
remains underexplored. Moreover, existing simplicity bias measures have primarily been applied to simple models (e.g., MLPs), and low-dimensional input space, without validation on large models.
We aim to fill these gaps by devising a proxy measure of simplicity that improves over existing ones, then examining how this measure correlates with various models' performance on various image-classification datasets.

In this paper, we focus specifically on CLIP models~\cite{radford2021learning} and fine-grained image classification tasks.
We consider recent methods that allow modulating the SB in a model
like smoothed activations~\cite{hu2025curvature} and LayerNorm scaling~\cite{Teney_2024_CVPR} then observe how this affects a model's performance on various tasks.
Technically, a key challenge lies in evaluating the SB of trained vision models, which are often very similar due to being trained on large-scale data. Differences across models can be subtle. We present a simplified theoretical case (Section~\ref{sec:analysis}) that defines model complexity and analyzes model outputs under perturbations, showing that existing measures based on output sensitivity cannot reliably distinguish a truly complex model from a simple one with large outputs. Moreover, previous methods consider only equidistance in the input space, neglecting the crucial influence of the spectral domain for image tasks.
Therefore, we propose a new measure that decouples the image space into several frequency bands and quantifies sensitivity of outputs on each band. Empirically, we find that when SB is strengthened by recent methods, the sensitively of high-frequency components  is consistently suppressed more strongly than that of low-frequency components, whereas the prior measure fail to capture meaningful trends. The phenomenon we find indicates that a simpler model prefers to learn low-frequency features.
Finally, we evaluate model performance across diverse image classification tasks under varying simplicity bias induced by our adapted versions of the methods in \cite{hu2025curvature,Teney_2024_CVPR} (see Section~\ref{sec:modulate}). Extensive results show that tasks such as OOD classification and adversarial robustness differ in their bias preference, highlighting the need for task-specific bias design.

Our contributions are summarized as follows.
\begin{itemize}[leftmargin=1em]

\item We introduce a mathematically tractable definition of model complexity based on the maximum derivative, and use a Taylor expansion to analyze the limitations of existing sensitivity-based SB measures (Section~\ref{sec:analysis}).

\item We propose a new frequency-aware measure to examine the SB of CLIP models, which captures subtle differences by two SB-modulation methods (Sections~\ref{sec:method}--\ref{sec:effect}) and reveals that the model tends to be sensitive to low-frequency features when SB is strengthened.

\item We leverage SB-modulation methods to investigate how simplicity bias correlates with the performance of CLIP models across a range of fine-grained image classification tasks,  including zero-shot classification, out-of-distribution (OOD) generalization, adversarial robustness, transfer adversarial robustness, and robustness to image corruptions.

\item We find strong links between a model's SB and downstream performance. Specifically:
(1) {For zero-shot classification}, varying SB helps mitigate the discrepancy between the model's inherent bias and the task characteristics. 
(2) For fine-tuning tasks (see Figure~\ref{fig:abstractfig}), models with higher complexity exhibit stronger robustness against adversarial attacks (e.g., PGD, MI-FGSM, BIM), whereas models with a stronger SB achieve superior performance in OOD generalization (i.e., CIFAR-10, iWildCam) and robustness under transfer attacks.
(3)  Comparison between architectures further supports the correlation between fine-tuning tasks and SB: We find that ViT generally requires less SB strengthening, or even a shift toward higher complexity, to achieve optimal performance on downstream tasks(Section~\ref{sec:ood},\ref{sec:attack}). This is consistent with its inherently stronger simplicity bias compared to ResNet (Section~\ref{sec:effect}).
\end{itemize}

\section{Related Work}

\subsection{Simplicity Bias and Generalization}
The ``inductive bias'' of a learning algorithm refers to the set of assumptions used to generalize from a limited set of training examples. Notably, SB is one form of inductive bias observed in modern neural networks, referring to the tendency to prefer simple functions even before training~\cite{mingard2019neural,Teney_2024_CVPR,valle2019deep}.
A large body of work links simplicity bias~\cite{valle2019deep,Teney_2024_CVPR,Teney_2025_CVPR,kalimeris2019sgd,teney2022evading,park2022vision,arpit2017closer} to the generalization capabilities of neural networks. For example, Valle-Pérez et al.\cite{valle2019deep} show that the parameter–function map of DNNs is biased towards low-complexity functions, which facilitates generalization. Teney et al.\cite{Teney_2025_CVPR} train various MLP models with different levels of simplicity bias and show that inducing a simplicity prior improves performance on image tasks.
In this paper, rather than simply concluding that SB benefits image tasks, we further explore and distinguish the role of SB across different types of image tasks.

\subsection{Measuring Simplicity Bias}
Previously, simplicity bias was measured via low-dimensional input projections~\cite{morwani2023simplicity}, decision boundary linearity~\cite{shah2020pitfalls}, or accuracy differences on spurious versus invariant features, all of which rely on carefully controlled datasets and model performance. More recently, several studies have examined the nonlinearity of a model's input-output mapping, either through Fourier analysis on evenly sampled inputs~\cite{Teney_2024_CVPR} or by evaluating output variation along interpolation paths~\cite{Teney_2025_CVPR}.Sensitivity-based methods are easier to apply and largely independent of performance, but they typically compare models on different tasks, making comparison of large models on the same task still challenging. We address this by proposing a frequency-based approach in Section~\ref{sec:method}.

\subsection{Simplicity Bias Modulation}
To better examine ``simplicity'' properly of networks or to improve performance, numerous modulation methods have been proposed based on different interpretations of simplicity bias, including activation function design~\cite{hu2025curvature,ziyin2020neural,Biswas_2022_CVPR,sitzmann2020implicit}, weight regularization~\cite{Teney_2024_CVPR,ShiIJCV22}, and data augmentation~\cite{yucel2023hybridaugment++}.
For example, periodic activations~\cite{ziyin2020neural,sitzmann2020implicit} have been adopted and shown superior performance over ReLU in physics-informed modeling and financial forecasting. 
Hu et al.~\cite{hu2025curvature} propose a ReLU variant with controllable smoothness, aiming to tune the curvature of pre-trained ResNets. 
Teney et al.~\cite{Teney_2024_CVPR} find that LayerNorm magnitudes can modulate the complexity of MLPs. However, these modulation methods have not been thoroughly examined with SB analysis tools on large models. In this paper, we adopt the approaches of~\cite{hu2025curvature} and~\cite{Teney_2024_CVPR} to assess their SB impact on ResNets and ViTs, respectively, given their architectural compatibility.

\section{Limitations of Existing Simplicity Bias Measures}
\label{sec:analysis}

In this section, we consider a simplified setting of a network with 1-D input and output. We first provide a mathematically tractable definition of model complexity, and then analyze model outputs under perturbations of varying complexity levels. This analysis reveals the limitations of existing simplicity bias measurements and inspired our design for frequency-aware simplicity bias analysis in more realistic settings (Section~\ref{subsec:mymethod}).


\subsection{How to Define Model Complexity?}

In order to quantify simplicity bias, we need to define model \emph{complexity}, with lower complexity indicating a stronger SB~\cite{Teney_2024_CVPR,Teney_2025_CVPR}. However, existing studies~\cite{Teney_2024_CVPR,Teney_2025_CVPR} directly use various sensitivity measurements of model output under uniformly spaced inputs as a proxy for complexity. Recent findings~\cite{sitzmann2020implicit,xu2019frequency} suggest that neural networks struggle to learn higher-order derivatives, implying that functions with significant high-order derivatives are inherently more complex. 
Motivated by this, we propose a simplified yet mathematically tractable definition of model complexity based on the relative magnitude of the highest-order derivatives, i.e.:


 \label{define} 
\begin{definition}[Model complexity]
\label{definition}
    Let a network \( f: \mathbb{R} \to \mathbb{R} \) be a real-valued function whose derivatives exist.
    The model complexity is then defined as


\begin{equation}
  \label{complexity}
\text{Complexity}(f) = \frac{M_d}{\sum_{k=0}^{d} M_k} \cdot d,
\end{equation}
\end{definition}
where \(M_k\) denotes the magnitude of the \(k\)-th order derivative, 
and \(d = \max \{ k \mid \exists x, \, f^{(k)}(x) \text{ exists} \}\) is the highest-order derivative of $f$. This definition considers both the highest-order derivative and its relative contribution.
 A constant function is the simplest with \( d=0 \). This definition roughly aligns with the intuition behind Kolmogorov complexity~\cite{valle2019deep}. 


\subsection{Model Output under Perturbations}
\label{perturb}
 Consider  a small perturbation $\delta(x)$ of the form
\begin{equation}
\delta(x) = \varepsilon \,\phi\bigl( x\bigr).
\label{eq:delta}
\end{equation}
Here, \(\varepsilon>0\) denotes the maximum perturbation amplitude. We assume that  \(\phi\) is a real‐valued function with \(\|\phi\|_\infty \leq 1\) and that \(\phi\) admits a well-defined Fourier transform. However, our analysis does not rely on the specific form of \(\phi\).  Consequently, \(\lvert \delta(x)\rvert \le \varepsilon\) for all \(x\).
In the following, we abbreviate \( \delta \) as \( \delta(x) \)  for brevity. 
We apply the Taylor expansion of \( f \) around \( x \) with the highest-order derivative $d$
\begin{equation}
f(x + \delta) = f(x) + f'(x) \delta + \frac{f''(x)}{2!} \delta^2 + \cdots + \frac{f^{(d)}(x)}{d!} \delta^d + R_d(\delta),
\label{eq:taylor}
\end{equation}
where  \( R_n(\delta) \) denotes the remainder term.
Therefore, the  change of model output is

\begin{equation}
\Delta f = f(x + \delta) - f(x) = f'(x) \delta +  \cdots + \frac{f^{(d)}(x)}{d!} \delta^d + R_d(\delta).
\label{eq:deltaf}
\end{equation}

Assuming the $k$-th derivative of $f$ is uniformly bounded over the interval $[x, x+\delta]$, 
\[
|f^{(k)}(t)| \leq M_k, \quad \forall t \in [x, x+\delta], \quad 1 \leq k \leq d.
\]

Taking the absolute value of the change in  \( f \), we have
\begin{align}
|\Delta f| &\leq |f'(x)| \cdot |\varepsilon| + \sum_{k=2}^d \frac{M_k}{k!} \cdot \varepsilon^k.\label{eq:deltaf} 
\end{align}
This equation shows the maximum local output change of \( f \) at a point \( x \) where the \( d \)-th derivative exists.

\noindent\textbf{Limitations.} Equation~\ref{eq:deltaf} reveals that, for a fixed perturbation amplitude \(\epsilon\), 
the local sensitivity of the model is determined not only by the order of derivatives \(d\) 
but also by all coefficients \(M_k\), which leads to two limitations: 
(1) the output changes cannot distinguish the contributions of low- or high-order derivative components, and 
(2) the output changes cannot differentiate between weak expressive power (i.e., \(M_k\) is small for all terms) and a tendency towards simple functions (i.e., the coefficients of low-order terms \(M_k\) are dominant).

This suggests that \textbf{recent simplicity bias measurements~\cite{Teney_2024_CVPR,Teney_2025_CVPR}, which rely on output changes (Section~\ref{subsec:Preliminary}), may only roughly reflect model complexity.} Although we consider only a local point and ignore derivatives elsewhere, we empirically validate this concern in Section~\ref{sec:effect}.

\begin{figure*}[t]
    \centering
    \begin{subfigure}[t]{0.98\textwidth}
        \centering
        \includegraphics[width=\textwidth]{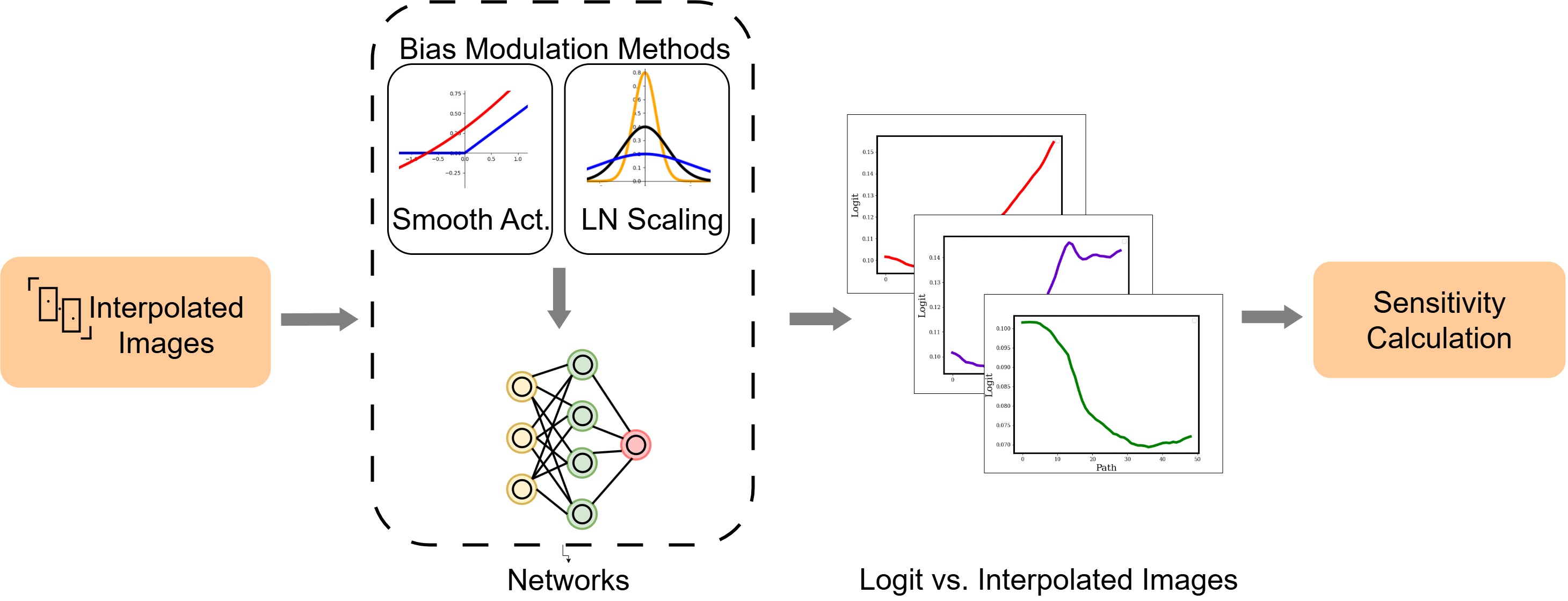}
        \caption{Measure pipeline. Interpolated images at different frequency bands are fed to networks adjusted by two SB-modulation methods, and SB is measured by logit sensitivity in each band.}
        \label{fig:sub1}
    \end{subfigure}
    \hfill
    \begin{subfigure}[t]{0.98\textwidth}
        \centering
        \includegraphics[width=\textwidth]{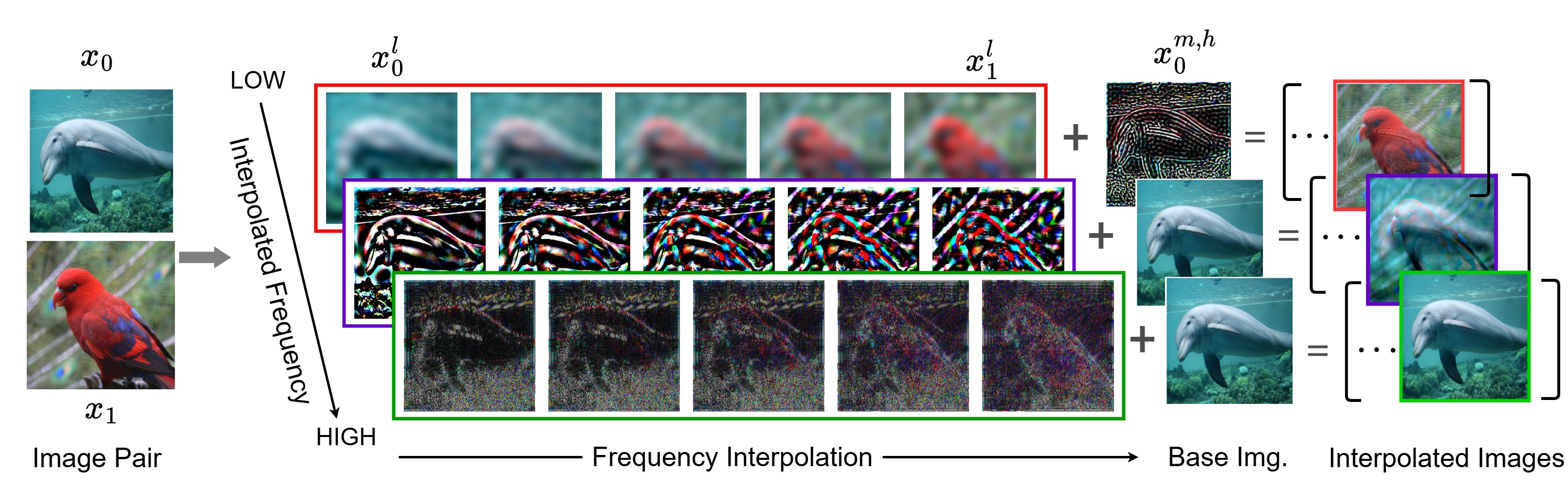}
        \caption{Interpolated image generation for each frequency. $x_0, x_1$ are two images sampled from different classes. $x^l,x^m,x^h$ denote low, middle, and high-frequency components, respectively. $x^{m,h}$ indicates that middle and high-frequency components are preserved in this image. Interpolation images are generated by blending base image into frequency interpolation. }
        \label{fig:sub2}
    \end{subfigure}
    \caption{Proposed frequency-aware measure of simplicity bias. ~\ref{fig:sub1} shows the main pipeline, and ~\ref{fig:sub2} illustrates the generation of interpolated images for the network.}
    \label{fig:pipeline}
\end{figure*}

\begin{figure*}[t!]
    \centering
    \begin{subfigure}[t]{0.98\linewidth}
        \centering
        \includegraphics[width=\linewidth]{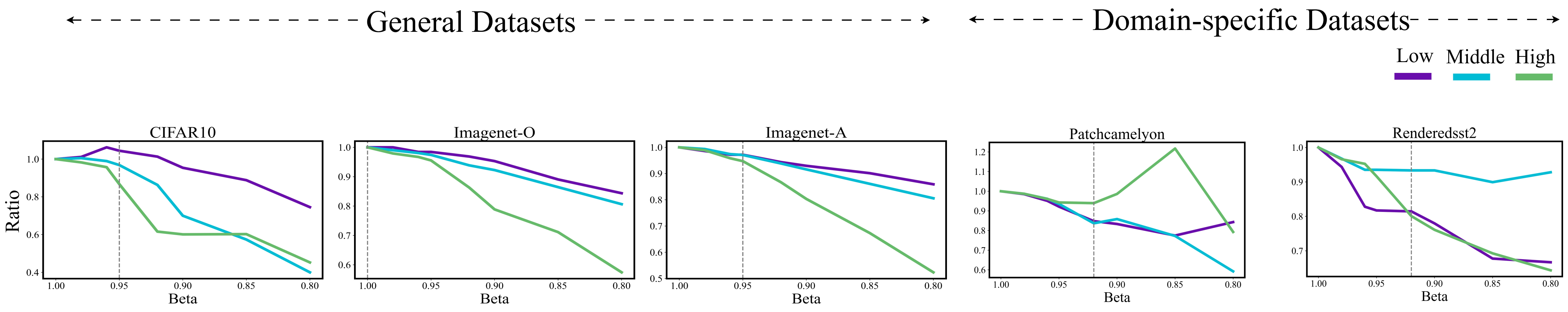}
        \caption{Ours: CLIP (ResNet-50)}
        \label{fig:lmd_complexity_pretrain}
    \end{subfigure}
      \vspace{1em}
    \begin{subfigure}[t]{0.98\linewidth}
        \centering
        \includegraphics[width=\linewidth]{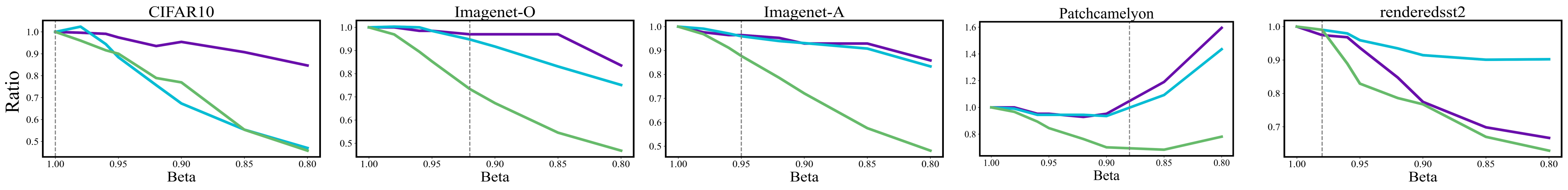}
        \caption{Ours: CLIP (ResNet-101)}
        \label{fig:lmd_complexity_pretrain2}
    \end{subfigure}
          \vspace{1em}
    \begin{subfigure}[t]{0.98\linewidth}
        \centering
        \includegraphics[width=\linewidth]{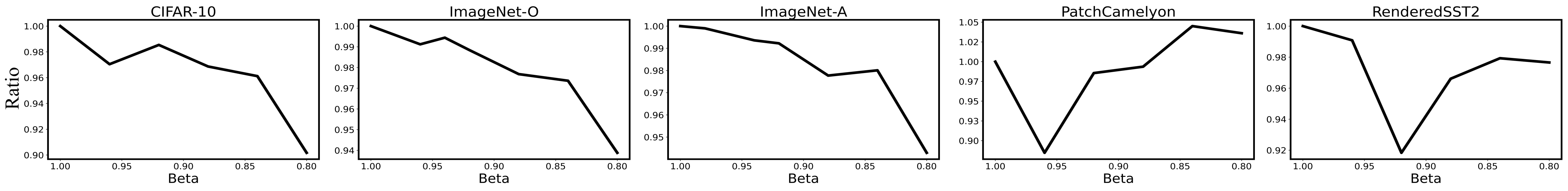}
        \caption{Baseline: CLIP (ResNet-101)}
        \label{fig:lmd_complexity_pretrain3}
    \end{subfigure}
    
    \caption{
         Sensitivity decay ratio of baseline method adapted from \cite{Teney_2025_CVPR} and our proposed frequency-wise metric with respect to $\beta$. Dash line indicates the optimal $\beta$ for each model. 
        In most datasets, smaller $\beta$ tends to preserve model sensitivity to low-frequency features while suppressing high-frequency ones. Our method is more informative and consistent compared to the baseline method.
    }
    \label{fig:lmd_complexity}
\end{figure*}
\section{How to Measure Simplicity Bias for Modern DNNs? }
\label{sec:method}
In this section, we present how to adapt the SB measurement \cite{Teney_2025_CVPR} to the CLIP model (referred to as the Baseline). and then propose our frequency-aware method.
\subsection{Preliminaries}
\label{subsec:Preliminary}

\paragraph{CLIP overview}
 We first define \( (x, {y})  \)  an  image and its corresponding label sampled from a classification dataset $\mathcal{D}$. Given an image encoder $f^{I}$ and a text encoder $f^{T}$, the logits of CLIP is denoted as
\begin{equation}
    \mathbf{z} =  f^{I}\left({x}\right)\cdot f^{T}(\mathcal{T}),  
\end{equation}
where $\mathcal{T}$  is a set of text prompts.
\paragraph{Baseline method for CLIPs.}

Given two images ${x}_1, {x}_2 \in \mathbb{R}^d$ from different classes, we generate a linearly interpolated path in the input space:
\begin{equation}
    \label{eq:path}
    {X}_{{x}_1,{x}_2} = \left[ (1-\lambda){x}_1 + \lambda {x}_2 \right] \in \mathbb{R}^{n \times d},
\end{equation}
where  $\lambda \in \left\{ 0, \frac{1}{n-1}, \frac{2}{n-1}, \dots, 1 \right\}$, $d$ is the dimensions of a image.   
The predicted logits along this interpolation path are computed as:
\begin{equation}
    {\mathbf{Z}} = f^{I}\left(\mathbf{X}_{{x}_1,{x}_1}\right)\cdot f^{T}(\mathcal{T}),
\end{equation}

We denote the  logit corresponding to label $y_i$ as  \( \mathbf{z} \). 
Sensitivity is measured by the smoothness of   \( \mathbf{z} \) using Total Variation (TV)~\cite{Teney_2025_CVPR} as the metric:
\begin{equation}
\label{eq:sen}
   \mathcal{S} =  \mathbb{E}_{{x}_1, {x}_2 \sim \mathcal{D}}\sum_{q=0} ^{n-2}\frac{1}{\|{x}_1-{x}_2\|_2}\left| {z}_{q+1} - {z}_{q} \right|,
\end{equation}
where \( \|\cdot\|_2 \) denotes L2 norm used to normalize the distance between images. We take $\mathcal{S}^{-1}$ as the level of simplicity bias, so that a larger value indicates a stronger bias toward simpler functions. We refer to this adopted measurement as the Baseline for the rest of the paper.

\subsection{Frequency-aware Simplicity Bias Measurement}
\label{subsec:mymethod}
As the model function $f$ and its derivatives are not directly accessible (see Equation~\ref{complexity}), we instead analyze the model's sensitivity to different frequency components of images. Low-frequency inputs correspond to gradual changes in natural images, so strong sensitivity to them reflects a preference for ``simple'' solutions. We decompose images into low-, mid-, and high-frequency components (LFC, MFC, HFC) using masked inverse Fourier transforms, and compare the model's sensitivity across them. This design avoids conflating simplicity bias with limited expressive capacity.
Formally, 
given a frequency band $r_k =[r_{\min}^{(k)}, r_{\max}^{(k)}],\ k \in \{l, m, h\}$, we first define a band-pass function $m(\cdot;r_k)$ whose value at frequency indices  $(i,j)$ is
\begin{equation}
{m}(x,r_k)_{i,j} = 
\begin{cases}
\mathcal{F}({x})_{i,j}, & \text{if }  r_{\text{min}}^{(k)} \leq \rho(i,j) < r_{\text{max}}^{(k)}, \\
0, & \text{otherwise},
\end{cases}
\end{equation}
where $\rho(i,j)$ denotes the Euclidean distance from the frequency index $(i,j)$ to the center of the frequency domain, and $\mathcal{F}({x})_{i,j}$ denotes the corresponding discrete Fourier transform coefficient of ${x}$.

Then, we denote the k-frequency components of ${x}$ as
\begin{equation}
    {{x}^k} = \mathcal{F}^{-1}({m}({x},r_k)),
\end{equation}

where $\mathcal{F}^{-1}$ is inverse Fourier transform.  The thresholds for the low-, mid-, and high-frequency bands are set based on the normalized distance from the center of the frequency domain, corresponding to 0–25\%, 25–80\%, and 80–100\% of the maximum frequency, respectively. 
To preserve all frequency components of inputs, only target frequency component is interpolated, rest of frequency components are intact. The interpolated path for $k$-frequency component between ${x}_1$ and $\mathbf{x}_2$ is then defined as 
\begin{equation}
\label{eq:fre_path}
\mathbf{X}^{k}_{{x}_1, {x}_2} = {x}_1 + \left[ (1 - \lambda)\, {x}_1^{k} + \lambda\, {x}_2^{k} \right], \quad \text{where } k \in \{l, m, h\}.
\end{equation}
Then, we obtain the logits along this path, denoted as
\begin{equation}
    {\mathbf{Z}^k} = f^{I}\left(\mathbf{X}^k_{x_{1},x_{2}}\right)\cdot f^{T}(\mathcal{T}).
\end{equation}
  The sensitivity for each frequency band  is measured as
\begin{equation}
\label{eq:sen}
   \mathcal{S}^ {k} =  \mathbb{E}_{{x}_1, {x}_2 \sim \mathcal{D}}\sum_{q=0} ^{n-2}\frac{1}{\|{x}_1^k-{x}_2^k\|_2}\left| {z}^{k}_{q+1} - {z}^{k}_{q} \right|,
\end{equation}
where $k\in \{l, m, h\}$. Finally, we quantify simplicity bias as the ratio of low-frequency sensitivity, i.e.,
\begin{equation}
  \label{eq:ratio}
\text{Ratio} = \frac{\mathcal{S}^{l}}{\mathcal{S}^{h}}.
\end{equation}
 A bigger value indicates stronger simplicity bias. By defining SB as the proportion of low-frequency sensitivity, we avoid conflating bias to simple function with under-expressivity. The measure is illustrated    in Figure~\ref{fig:pipeline}.

\begin{figure*}
    \centering
    \includegraphics[width=0.95\linewidth]{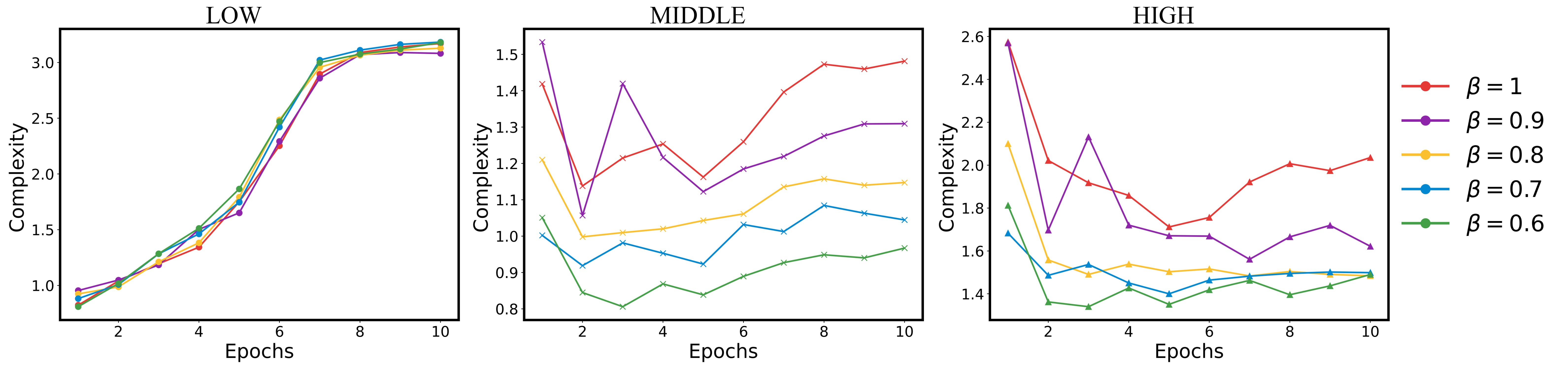}
    \caption{Sensitivity to low-, mid-, and high-frequency components during CIFAR-10 fine-tuning. Each curve represents CLIP (ResNet-50) with a different $\beta$. Models with stronger SB retain sensitivity to low-frequency features while suppressing sensitivity to high-frequency features.}
    \label{fig:comlex_ft}
\end{figure*}
\section{Simplicity Bias Modulations for CLIPs}
\label{sec:modulate}
To investigate the role of simplicity bias, we require modulation methods capable of adjusting simplicity bias. Here, we present two inductive bias modulation approaches based on the concept of ``simplicity'',  which theoretically modulate simplicity bias but lack of understanding on various settings on CLIP mdoels.

For ResNet-based encoders, we replace all ReLU layers in the vision backbone with BetaReLU~\cite{hu2025curvature}, which introduces a smoothness-controlling parameter $\beta$ to ReLU. The activation function is defined as
\begin{equation}
f(x) = 0.5\, \sigma\left(\frac{\beta x}{1 - \beta}\right) \cdot x 
+ 0.5 \log\left(1 + e^{\frac{x}{1 - \beta}}\right) \cdot (1 - \beta),
\end{equation}
where $\sigma(\cdot)$ denotes the sigmoid function. Setting $\beta = 1$ recovers the standard ReLU, while smaller values of $\beta$ yield smoother activations. 
While BetaReLU has shown benefits for pre-trained ResNets in a training-free setting~\cite{hu2025curvature}, its effect on CLIP models and fine-tuning settings remain inconclusive.

For ViT-based encoders~\cite{dosovitskiy2020image}, BetaReLU is not applicable. Inspired by the MLP behavior in \cite{Teney_2024_CVPR}, we add an extra scalar $\gamma_s$ to the original learnable weight $\gamma$ of each LayerNorm layer in the vision encoder, i.e.,

\begin{equation}
\text{LayerNorm}(x) = \left( \frac{x - \mu}{\sigma} \right) \cdot (\gamma \cdot \gamma_s) + \beta.
\end{equation}
We refer to this method as \emph{LayerNorm scaling}. To avoid excessively restricting model capacity, 
 $\gamma_s$ should not be set too small.  In all fine-tuning experiments, we set $\gamma_s \in [0.9, 1.5]$. A larger $\gamma_s$ corresponds to higher model complexity compared to the original pre-trained weights.

\begin{table}[t]
\centering
\caption{
Quantification of simplicity bias in fine-tuned CLIP models. “Sen. Low” denotes sensitivity to LFC. The sensitivity ratio (Eq.~\ref{eq:ratio}) serves as a proxy for simplicity bias. Decreasing $\beta$ or $\gamma_s$ consistently shifts the model’s bias toward lower frequencies. Baseline~\cite{Teney_2025_CVPR} is included for comparison, showing no consistent trend.
For comparison, the baseline is also reported but does not exhibit a consistent trend.
}
\label{tab:freq_sens}
\vspace{0.3em}
\renewcommand{\arraystretch}{1.2}
\setlength{\tabcolsep}{6pt}
\resizebox{\linewidth}{!}{%
\begin{tabular}{ll|ccccc}
\toprule
\multicolumn{7}{c}{\textbf{ResNet-CLIP (varying $\beta$)}} \\
\midrule
Backbone & Metric & $\beta=1.0$ & 0.9 & 0.8 & 0.7 &0.6\\
\cmidrule(lr){1-2} \cmidrule(lr){3-7}
\multirow{4}{*}{RN50}
  & baseline~\cite{Teney_2025_CVPR}    & 0.31 & 0.31 & 0.32 & 0.31 & 0.31 \\
  & Sen. Low   & 3.28 & 3.26 & 3.25 & 3.26 &3.22\\
  & Sen. High  & 2.06 & 1.62 & 1.45 & 1.36 &1.33\\
  & Ratio      & 1.59 & 2.01 & 2.24 & 2.39 &2.42\\
\cmidrule(lr){1-2} \cmidrule(lr){3-7}
\multirow{4}{*}{RN101}
  & baseline     & 0.23 & 0.22 & 0.22 & 0.24 & 0.24 \\
  & Sen. Low   & 4.24 & 4.36 & 4.43 & 4.0  & 4.03 \\
  & Sen. High  & 7.45 & 6.86 & 7.69 & 4.68 & 3.00 \\
  & Ratio      & 0.57 & 0.64 & 0.58 & 0.85 & 1.34\\
\midrule
\multicolumn{7}{c}{\textbf{ViT-CLIP (varying $\gamma_s$)}} \\
\midrule
Backbone & Metric & $\gamma_s=1.5$ & 1.2 & 1.1 & 1 & 0.9 \\
\cmidrule(lr){1-2} \cmidrule(lr){3-7}
\multirow{4}{*}{ViT-B/16}
  & baseline    & 0.28 & 0.33 & 0.31 & 0.32 & 0.32 \\
  & Sen. Low   & 3.11 & 2.60 & 2.61 & 2.64 & 2.55  \\
  & Sen. High  & 1.18 & 0.94 & 0.88 & 0.83 & 0.77  \\
  & Ratio      & 2.64 & 2.77 & 2.97 & 3.18 & 3.31  \\
\bottomrule
\end{tabular}
}
\end{table}

\section{Measuring Simplicity Bias in Practice}
\label{sec:effect}

We evaluate how the modulation methods described in Section~\ref{sec:modulate} affect simplicity bias in both pre-trained and fine-tuned CLIP models. We use the official pre-trained CLIP models released by OpenAI~\cite{cherti2023reproducible} for both zero-shot and fine-tuning experiments. All SB measurements are averaged over three runs, each based on 2000 image pairs.
\begin{table*}[h]
\centering
\setlength{\tabcolsep}{4pt}
  \caption{
    Zero-shot classification accuracy of $\text{CLIP}_{\text{RN}}$ with original ReLU and optimal BetaReLU activations. The optimal parameter is denoted by ``$\beta$''. The highest accuracy is highlighted in bold.  Optimal $\beta$ and effectiveness vary by dataset and model.
  }
  \label{table:zero}
\begin{tabular}{c|ccc|ccc}
\toprule
\multirow{2}{*}{Dataset} & \multicolumn{3}{c|}{RN50} & \multicolumn{3}{c}{RN101} \\
\cmidrule(lr){2-4} \cmidrule(lr){5-7}
 & ReLU & BetaReLU & $\beta$ & ReLU & BetaReLU & $\beta$ \\
\midrule
PatchCamelyon~\cite{veeling2018rotation} & 66.4 & \textbf{67.1} & 0.92 & 56.1 & \textbf{61.5} & 0.88 \\
DTD~\cite{cimpoi14describing}            & 39.8 & \textbf{40.1} & 0.91 & 40.1 & \textbf{40.7} & 0.94 \\
ImageNet-O~\cite{hendrycks2021natural}     & 56.5 & 56.5 & 1 & 50.1 & \textbf{50.4} & 0.94 \\
ImageNet-A~\cite{hendrycks2021natural}     & 23.9 & \textbf{24.0} & 0.98 & 31.0 & \textbf{31.4} & 0.94 \\
CIFAR10~\cite{krizhevsky2009learning}        & 70.4 & \textbf{73.5} & 0.94 & 78.9 & 78.9 & 1 \\
Pets~\cite{parkhi2012cats}           & 85.7 & 85.7 & 1 & 86.9 & \textbf{87.0} & 0.98 \\
Rendered SST2~\cite{radford2021learning}  & 55.8 & \textbf{56.6} & 0.92 & 64.3 & \textbf{64.6} & 0.98 \\
CLEVR-All~\cite{johnson2017clevr}      & 21.5 & 21.5 & 1 & 24.1 & \textbf{26.2} & 0.96 \\
\bottomrule
\end{tabular}
\end{table*}
\subsection{Implementation Details}
\label{basicsetting}
\noindent\textbf{Pre-trained model settings.}
We choose ResNet-50 and ResNet-101 as vision backbones due to their similar model sizes. ViT-based backbones are excluded since BetaReLU does not support ViT-B/16 architectures.
 For clarity, we \textbf{refer to each CLIP model by its backbone name} (i.e., RN50, RN101, and ViT16) throughout the paper. We evaluate on the following datasets: ImageNet-O~\cite{hendrycks2021natural}, ImageNet-A~\cite{hendrycks2021natural}, CIFAR-10~\cite{krizhevsky2009learning}, Rendered SST2~\cite{radford2021learning}, and PatchCamelyon~\cite{veeling2018rotation}. The latter two are domain-specific datasets for sentiment analysis and medical diagnosis. \\
\noindent\textbf{Fine-tuning settings.}
We adopt pre-trained ResNet-50, ResNet-101, and ViT-B/16 for fine-tuning.  Fine-tuning follows the protocol of~\cite{wortsman2022robust}, where the entire image encoder is updated jointly with a linear classifier initialized from the  weights of the corresponding text encoder. Each model is fine-tuned for 10 epochs using a learning rate of $1 \times 10^{-5}$ and a batch size of 64. We train and evaluate models on CIFAR-10 dataset.\\

\subsection{Experiments}
\paragraph{On pre-trained models.}
Figure~\ref{fig:lmd_complexity} shows the decay ratio of sensitivity on LFC, MFC, and HFC with respect to the parameter $\beta$. The ratio is computed by dividing the sensitivity at each $\beta$ by that of the original model ($\beta{=}1$). Smoother activation functions (smaller $\beta$) lead to a much faster decay of sensitivity on HFC than on LFC, indicating an increased bias toward low-frequency features. Some exceptions are observed on domain-specific datasets that significantly deviate from the pre-training distribution.
\paragraph{On fine-tuned models.}
Table~\ref{tab:freq_sens} reports our frequency-aware metric and the baseline metric on fine-tuned models. For ResNets, the low-frequency ratio increases consistently as $\beta$ decreases, consistent with trends observed on pre-trained models. A spike in high-frequency sensitivity at $\beta = 0.8$ for RN101 may stem from numerical instability in metric computation.

For ViT-B/16, increasing the initialized LayerNorm scaling factor $\gamma_s$ consistently reduces the low-frequency ratio, demonstrating effective control of simplicity bias through $\gamma_s$. Moreover, ViT exhibits a substantially stronger low-frequency preference compared to both ResNet variants.

In contrast, the baseline metric fails to capture these phenomenons under both ResNet and ViT architectures, showing neither consistent trends across $\beta$ nor $\gamma_s$. This is because  low-frequency sensitivity dominates the  model's overall response, thereby limiting further analysis.
\paragraph{On fine-tuning dynamics.}
Figure~\ref{fig:comlex_ft} further shows the fine-tuning dynamics of RN50. A smaller $\beta$ mitigates the increase in mid- and high-frequency sensitivity, while maintaining strong low-frequency ones. Interestingly, a U-shaped sensitivity trend is observed on mid-frequency components, and on high-frequency components when $\beta=1$. This phenomenon indicates that the model first mitigates the distribution mismatch between pre-training and fine-tuning data at these frequencies by decreasing sensitivity, and then begins to overfit to mid- or high-frequency functions. This provides a new perspective for understanding fine-tuning dynamics.
\paragraph{Implications.} 
Our findings suggest that models with stronger SB tend to be less sensitive to high-frequency components. This aligns with the intuition that, in natural images, relying on high-frequency features requires representing functions with rapid variations, corresponding to more complex solutions that go against the simplicity bias.

\section{Simplicity Bias for Zero-shot Classification}%
\label{sec:pre}

\begin{table*}[t]
  \centering
  \caption{
Evaluation of RN50 and RN101 across low-, middle-, and high-frequency components. We report Sensitivity (Sen.), Accuracy Contribution (Acc. Contr.), and Accuracy Gain ($\Delta$Acc) under optimal $\beta$.  BetaReLU primarily improves RN50 via low-frequency components, and RN101 via high-frequency components. * denotes no $\beta$ leads to improvement over the baseline.
  }
  \label{tab:complexity-zeroshot}
  \vspace{0.5em}

  \begin{subtable}[t]{\linewidth}
    \centering
    \caption{RN50 results}
    \label{tab:sdr50}
    \vspace{0.3em}
    \resizebox{\textwidth}{!}{%
\begin{tabular}{c|ccc|ccc|ccc}
    \toprule
    \multirow{2}{*}{Dataset} & \multicolumn{3}{c|}{Low} & \multicolumn{3}{c|}{Middle} & \multicolumn{3}{c}{High} \\
    \cmidrule(lr){2-4} \cmidrule(lr){5-7} \cmidrule(lr){8-10}
    & Sen.  & Acc Contr. & $\Delta$Acc & Sen.  & Acc Contr. & $\Delta$Acc & Sen.  & Acc Contr. & $\Delta$Acc \\
    \midrule
    CIFAR10         & 0.0609 & 62.02 & \textbf{7.1} & 0.0190 & 6.94 & -3.4 & 0.0747 & 1.37 & -0.8 \\
    RenderedSST2    & 0.0372 & 50.08 & 0 & 0.0556 & 5.93 & -1.6 & 0.0527 & -0.27 & \textbf{2.4} \\
    ImageNet-A      & 0.0071 & 4.44  & 0 & 0.0309 & 12.68 & \textbf{0.4} & 0.0361 & 6.77 & 0 \\
    ImageNet-O*     & 0.0064 & 35.20 & 0 & 0.0310 & 19.50 & 0 & 0.0336 & 1.65 & 0 \\
    PatchCamelyon   & 0.0067 & 55.49 & \textbf{14.6} & 0.0233 & 11.77 & -13.2 & 0.0573 & -0.93 & 2.7 \\
    \midrule
    \textbf{Average} & 0.0237 & 41.45 & \textbf{3.4} & 0.0320 & 11.76 & -3.56 & 0.0509 & 1.72 & 0.86 \\
    \bottomrule
\end{tabular}
    }
  \end{subtable}

  \vspace{1em}

  \begin{subtable}[t]{\linewidth}
    \centering
    \caption{RN101 results}
    \label{tab:sdr101}
    \vspace{0.3em}
    \resizebox{\textwidth}{!}{%
\begin{tabular}{c|ccc|ccc|ccc}
\toprule
\multirow{2}{*}{Dataset} & \multicolumn{3}{c|}{Low} & \multicolumn{3}{c|}{Middle} & \multicolumn{3}{c}{High} \\
\cmidrule(lr){2-4} \cmidrule(lr){5-7} \cmidrule(lr){8-10}
& Sen. & Acc. Contr. & $\Delta$Acc  & Sen. & Acc. Contr. & $\Delta$Acc  & Sen. & Acc. Contr. & $\Delta$Acc  \\
\midrule
CIFAR10*         & 0.0789 & 74.70 & 0 & 0.0251 & 4.12  & 0 & 0.0703 & 0.04 & 0 \\
RenderedSST2     & 0.0315 & 50.03 & 0 & 0.0737 & 10.93 & -0.3 & 0.0533 & 3.29 & \textbf{0.6} \\
ImageNet-A       & 0.0085 & 6.59  & -0.3 & 0.0349 & 18.40 & \textbf{0.5} & 0.0351 & 6.01 & 0.2 \\
ImageNet-O       & 0.0067 & 36.95 & -3.4 & 0.0326 & 13.05 & \textbf{2.8} & 0.0361 & 0.00 & 0.9 \\
PatchCamelyon    & 0.0042 & 53.02 & 1.7 & 0.0108 & 3.92  & 1.0 & 0.0452 & -0.86 & \textbf{2.4} \\
\midrule
\textbf{Average} & 0.0260 & 44.26 & -0.4 & 0.0354 & 10.08 & 0.8 & 0.0480 & 1.70 & \textbf{0.82} \\
\bottomrule
\end{tabular}

  }
  \end{subtable}
\end{table*}

We investigate  the relationship between classification accuracy and simplicity bias on a range of zero-shot classification datasets. We begin by varying simplicity bias using BetaReLU and evaluating its effectiveness, followed by an in-depth analysis of the underlying mechanisms.

\subsection{Implementation Details}

\noindent\textbf{Models.}
All model configurations are consistent with pre-trained settings detailed in Section~\ref{basicsetting}.  

\noindent\textbf{Datasets.}
For zero-shot classification, we evaluate on the following datasets: ImageNet-O, ImageNet-A, CIFAR-10, Rendered SST2, and PatchCamelyon. Pets~\cite{parkhi2012cats}, Clevr-All~\cite{johnson2017clevr}, and DTD~\cite{cimpoi14describing} are used only in Section~\ref{sec:varies}. Our implementation is based on the CLIP-Benchmark codebase~\footnote{\url{https://github.com/LAION-AI/CLIP_benchmark}}.


\begin{figure}[h!]
  \centering
  \includegraphics[width=0.95\linewidth]{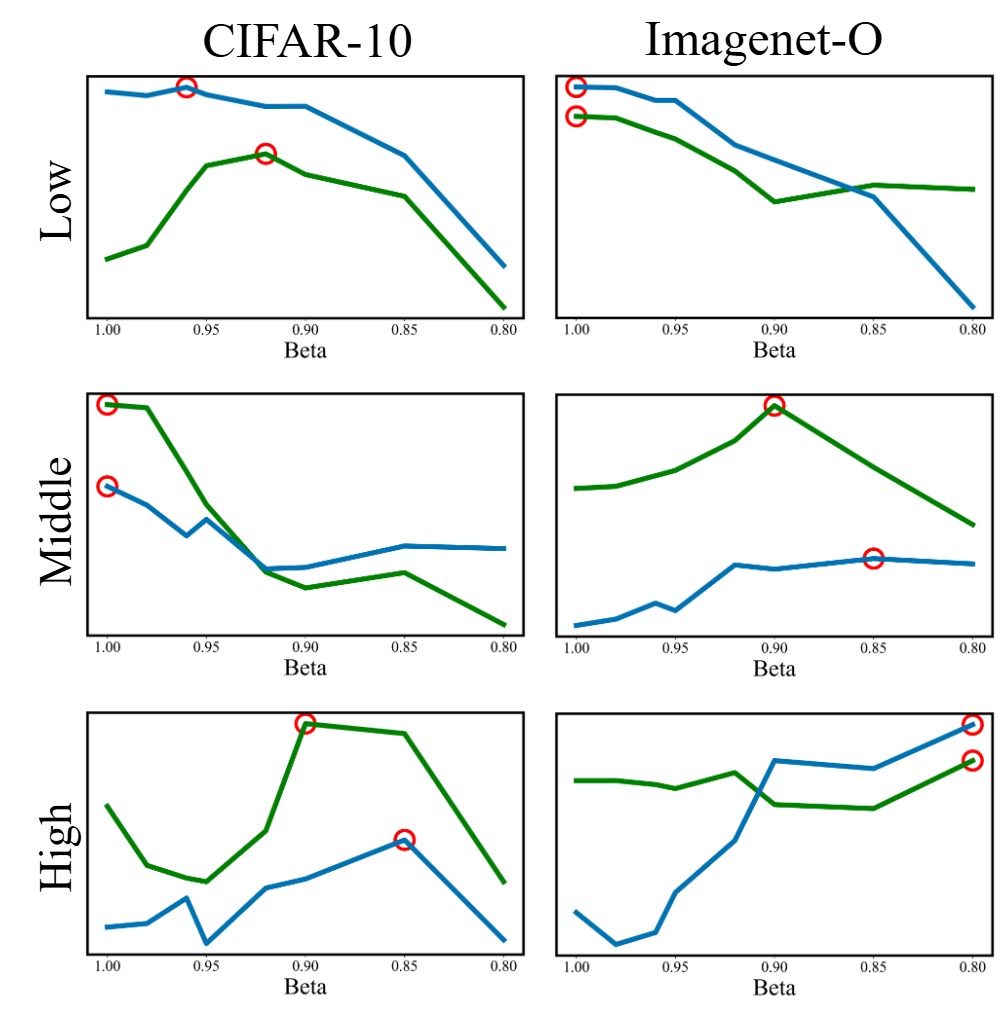}
  \caption{Accuracy contribution of RN50 (green) and RN101 (blue) on each frequency band. The red circle shows the highest contribution across $\beta$.  The dataset exhibits a similar distribution shift across models.}
  \label{fig:acccontri}
\end{figure}


\subsection{Simplicity Bias Modulation Improves Zero-shot Classification}
\label{sec:varies}
Table~\ref{table:zero} presents the zero-shot classification accuracy achieved by ReLU, the optimal BetaReLU, and the corresponding $\beta$. Despite the potential mismatch introduced by modifying the activation function without fine-tuning the pre-trained weights, BetaReLU nevertheless improves performance on most datasets. It drives us to further explore the underlying mechanism using our frequency-aware metric.

\subsection{Understanding Simplicity Bias Modulation via Frequency-aware Analysis}
\label{sec:indepth}
To investigate the observations in Section~\ref{sec:varies}, we analyze both the original and optimal BetaReLU models across low-, mid-, and high-frequency components.
Table~\ref{tab:complexity-zeroshot} reports three frequency-specific metrics: sensitivity, accuracy contribution (i.e., the accuracy gain obtained by cumulatively adding each frequency component, see Equation~\ref{ac}), and per-band accuracy gain ($\Delta$Acc) under the optimal~$\beta$ (see Equation~\ref{deltaacc}).  
Specifically, let $x^k$ be the $k$-frequency band of the image, $\mathrm{Acc}(\cdot)$ the classification accuracy, and $f$ / $\hat{f}$ the models with $\beta=1$ / optimal $\beta$. Take the mid-frequency band as example, we define
\begin{align}
\mathrm{AccContr}(f_m) &= \mathrm{Acc}\big(f(x^{l+m})\big) - \mathrm{Acc}\big(f(x^l)\big), \label{ac} \\
\Delta\mathrm{Acc}(m) &= \mathrm{AccContr}(\hat{f}_m)- \mathrm{AccContr}(f_m), \label{deltaacc}
\end{align}
where $x^{l+m}$ denotes the input containing only the low- and middle-frequency components. The insights are as follows:\\

\noindent\textbf{Deeper pre-trained models exhibit stronger simplicity bias.} As shown in Table~\ref{tab:complexity-zeroshot}, RN101 exhibits higher average sensitivity to low- and mid-frequency components than RN50 during pre-training, while its high-frequency sensitivity is relatively low. Although fine-tuning (Section~\ref{sec:effect}) increases RN101’s high-frequency sensitivity beyond RN50, its low-frequency sensitivity is still higher than RN50, which may contribute to better generalization on most natural tasks. We speculate that this stronger low-frequency bias in deeper models arises from the contrastive learning objective and large-scale pre-training, reinforcing the preference for low-frequency patterns. \\

\noindent\textbf{Simplicity bias modulation mitigates the suboptimal bias of models for downstream datasets.} Table~\ref{tab:complexity-zeroshot} indicates that models often exhibit suboptimal simplicity bias on certain datasets, with optimal solutions sometimes trending opposite to their inherent biases. For example, RN101, which has a strong inherent simplicity bias, benefits most from mid- or high-frequency components (0.82 or 0.80 vs. –0.40), whereas RN50, with a weaker simplicity bias, gains primarily from the low-frequency component (3.40 vs. –3.56 or 0.86).  

To further examine dataset influence, we plot how accuracy contribution changes with $\beta$ for each frequency band (Figure~\ref{fig:acccontri}). For RN101 on CIFAR-10, reducing $\beta$ improves low-frequency component performance but worsens mid-frequency performance, leaving the original setting ($\beta=1$) optimal overall. This explains the different behaviors of RN50 and RN101 on CIFAR-10. Moreover, the two models show consistent suboptimality or optimality across frequency bands on each dataset, reflecting distribution shifts between pre-training and fine-tuning data. For example, the mid-frequency band of CIFAR-10 and the low-frequency band of ImageNet-O already achieve optimal performance at $\beta=1$ for both models, while other frequencies remain suboptimal. These results highlight that optimal simplicity bias depends on both the downstream dataset and the model.

\paragraph{Model sensitivity vs. performance.}
\label{sensitivity-performance}
Table~\ref{tab:complexity-zeroshot} shows a positive correlation between model sensitivity to LFC and classification accuracy. Models with higher LFC sensitivity consistently perform better across all five datasets. No clear trend is observed for mid- or high-frequency components, indicating that LFC features transfer more effectively from pre-training to downstream tasks. Notably, RN50, though generally less sensitive to LFC, shows higher sensitivity on domain-specific datasets such as PatchCamelyon and RenderedSST2, suggesting it captures less common low-frequency patterns that remain informative OOD.

\section{Simplicity Bias in OOD Generalization }
\label{sec:ood}
In this section, we study how simplicity bias of fine-tuned models affects generalization under  distribution shift. Specifically, we fine-tune CLIP models by adjusting the simplicity bias—using $\beta$ for ResNets and the LayerNorm scaling factor $\gamma_s$ for ViTs—and evaluate their performance on Out-of-distribution benchmarks. 
%

\subsection{Implementation Details}

\paragraph{Training setup.}
All model configurations are consistent with the fine-tuning settings detailed in Section~\ref{basicsetting}.  Moreover, we fine-tune models on CIFAR-10 and iWildCam~\cite{beery2020iwildcam} under the same protocol.\\
\noindent\textbf{Datasets.}
The CIFAR-10 OOD benchmark includes two domain-shifted variants, CIFAR-10.1~\cite{recht2018cifar} and CIFAR-10.2~\cite{lu2020harder}, where we report top-1 accuracy. For iWildCam, we report both in-distribution and out-of-distribution micro-F1 scores.

\subsection{Main Findings}
Results for ResNet-based models on CIFAR-10 and iWildCam are reported in Tables~\ref{tab:ft-cifar} and~\ref{tab:ft-iwildcam}, respectively. Table~\ref{tab:vit_ft} shows the results for ViT-B/16 on both datasets.
\paragraph{Stronger simplicity bias improves performance on ID and OOD data.}
Tables~\ref{tab:ft-cifar}, \ref{tab:ft-iwildcam}, and \ref{tab:vit_ft} show that stronger simplicity bias can consistently improve performance across both in-distribution (ID) and out-of-distribution (OOD) tasks.  For example, on the iWildCam benchmark, RN50, RN101, and ViT-B/16 achieve relative   improvements of 4.8\%, 9.9\%, and 4.8\% on average accuracy, respectively. On CIFAR-10, relative improvements on average accuracy are 1.4\%, 0.96\%, 0.33\%, respectively.
\paragraph{Different OOD tasks exhibit different needs for simplicity bias.}
We observe that the optimal simplicity bias varies across OOD tasks. In the CIFAR-10 benchmark, the optimal values are $\beta = 0.7$ for both RN50 and RN101, and $\gamma_s = 0.9$ for ViT-B/16. In contrast, the iWildCam benchmark favors larger values: $\beta = 0.9$ for ResNets and $\gamma_s = 1.1$ for ViT-B/16. This suggests that CIFAR-10 benefits more from low-frequency inductive bias than iWildCam.

We speculate that this difference may arise from the distinct frequency bands that contain critical features in the two datasets. CIFAR-10 contains extremely low-resolution images ($32 \times 32$), where most semantic content is likely concentrated in low-frequency components~\cite{rahaman2019spectral}. In contrast, iWildCam is more ``complex'', featuring higher-resolution images and intricate patterns that require models to retain greater sensitivity to high-frequency components for effective discrimination. Our observations also suggest that the optimal simplicity bias is not universal, but depends on task-specific characteristics, which aligns with prior findings in~\cite{Teney_2025_CVPR}. 
\begin{table}[t]
\centering
\caption{
Accuracy (\%) of fine-tuned ResNet-based CLIP models on the CIFAR OOD benchmark.  The best average performance is achieved at $\beta = 0.7$ for both models.
}
\label{tab:ft-cifar}
\vspace{0.3em}
\resizebox{\linewidth}{!}{%
\begin{tabular}{c|ccc|c||ccc|c}
\toprule
\multirow{2}{*}{$\beta$} 
& \multicolumn{3}{c|}{RN50} & \multirow{2}{*}{Avg} 
& \multicolumn{3}{c|}{RN101} & \multirow{2}{*}{Avg} \\
\cmidrule(lr){2-4} \cmidrule(lr){6-8}
& CIFAR10 & 10.1~\cite{recht2018cifar} & 10.2~\cite{lu2020harder} &  & CIFAR10 & 10.1 & 10.2 &  \\
\midrule
1.0  & 96.01 & 90.20 & 86.00 & 90.74 & 96.87 & 91.60 & 86.90 & 91.79 \\
0.9  & 96.23 & 90.90 & 86.00 & 91.04 & 97.15 & 91.70 & 87.00 & 91.95 \\
0.8  & 96.36 & 91.35 & \textbf{87.65} & 91.79 & \textbf{97.38} & \textbf{92.60} & 87.60 & 92.53 \\
0.7  & \textbf{96.69} & \textbf{92.30} & 87.15 & \textbf{92.05} & 97.02 & 92.30 & \textbf{88.70} & \textbf{92.67} \\
0.6  & 96.58 & 91.15 & 87.40 & 91.71 & 97.03 & 92.40 & 87.65 & 92.36 \\
\bottomrule
\end{tabular}
}
\end{table}

\begin{table}[h]
\centering
\caption{
Macro-F1 (\%) of fine-tuned ResNet-based CLIPs on the iWildCam  benchmark. Compared to CIFAR, iWildCam shows a weaker preference for smooth activations. The best average performance is achieved at $\beta = 0.9$ for both models. 
}
\label{tab:ft-iwildcam}
\vspace{0.3em}
\begin{tabular}{c|cc|c||cc|c}
\toprule
\multirow{2}{*}{$\beta$} 
& \multicolumn{2}{c|}{RN50} & \multirow{2}{*}{Avg} 
& \multicolumn{2}{c|}{RN101} & \multirow{2}{*}{Avg} \\
\cmidrule(lr){2-3} \cmidrule(lr){5-6}
& ID & OOD & & ID & OOD & \\
\midrule
1.0 & 42.47 & 27.41 & 34.94 & 44.39 & 27.47 & 35.93 \\
0.9 & \textbf{46.20} & 27.01 & \textbf{36.61} & \textbf{48.02} & \textbf{30.93} & \textbf{39.48} \\
0.8 & 44.95 & \textbf{28.02} & 36.49 & 47.49 & 27.84 & 37.67 \\
\bottomrule
\end{tabular}
\end{table}
\begin{table}[h!]
\centering
\caption{
Fine-tuned ViT-B/16-CLIP performance on CIFAR and iWildCam benchmarks. The best average performance occurs at $\beta = 0.9$ for CIFAR and $\beta = 1.1$ for iWildCam, suggesting that ViT requires higher complexity than ReNet.
}

\label{tab:vit_ft}
\resizebox{\linewidth}{!}{%
\vspace{0.3em}
\begin{tabular}{c|ccc|c|cc|c}
\toprule
\multirow{2}{*}{$\gamma$} 
& \multicolumn{3}{c|}{\textbf{CIFAR (Accuracy)}} & \multirow{2}{*}{Avg} 
& \multicolumn{2}{c|}{\textbf{iWildCam (Macro-F1)}} & \multirow{2}{*}{Avg} \\
\cmidrule(lr){2-4} \cmidrule(lr){6-7}
& CIFAR10 & 10.1 & 10.2 & 
& ID & OOD & \\
\midrule
1.2 & 97.85 & 93.90 & \textbf{89.30} & 93.68 & 45.28 & \textbf{26.63} & 35.96 \\
1.1 & 97.99 & 93.90 & 88.90 & 93.60 & \textbf{46.90} & 26.55 & \textbf{36.73} \\
1.0 & \textbf{98.12} & 93.90 & 89.05 & 93.69 & 45.84 & 24.70 & 35.27 \\
0.9 & 98.04 & \textbf{94.70} & 89.25 & \textbf{93.99} & 44.29 & \textbf{26.63} & 35.46 \\
\bottomrule
\end{tabular}
}
\end{table}




\begin{table*}[h]
\centering
\caption{
Accuracy (\%) and clean accuracy of models fine-tuned with different $\beta$. Gradient-based and score-based attacks exhibit different preferences for simplicity bias.
}

\label{tab:adv_rn}
\renewcommand{\arraystretch}{1.2}
\resizebox{\linewidth}{!}{%
\begin{tabular}{c c c c c c|c|c c c c c c}
\toprule
\multirow{3}{*}{$\beta$} 
& \multicolumn{6}{c|}{RN50} 
& \multicolumn{6}{c}{RN101} \\
\cmidrule(lr){2-7} \cmidrule(lr){8-13}
& Clean & \multicolumn{4}{c|}{Gradient-based} & Score-based 
& Clean & \multicolumn{4}{c}{Gradient-based} & Score-based \\
 \cmidrule(lr){3-6} \cmidrule(lr){7-7} \cmidrule(lr){9-12} \cmidrule(lr){13-13}
& & PGD~\cite{madry2017towards} & FGSM~\cite{goodfellow2014explaining} & BIM~\cite{kurakin2018adversarial}  & MIFGSM~\cite{croce2020minimally} & Pixle~\cite{pomponi2022pixle} 
& & PGD & FGSM & BIM & MIFGSM & Pixle \\
\midrule
1.0 & 96.01 & \textbf{41.21} & \textbf{53.90} & \textbf{88.29} & \textbf{27.52} & 15.56 
    & 96.87 & 33.94 & 55.96 & 90.26 & {29.50}  &17.88  \\
0.9 & 96.23 & 40.68 & 52.34 & 86.77 & 23.23 & \textbf{17.78} 
    & 97.15 & \textbf{35.51} & \textbf{58.38} & \textbf{91.52} & \textbf{30.12} & 15.20 \\
0.8 & 96.36 & 28.07 & 50.13 & 86.97 & 20.90 & 12.83 
    & \textbf{97.38} & 22.43 & 52.22 & 87.58 & 19.82 & \textbf{18.13} \\
0.7 & \textbf{96.69} & 26.59 & 49.28 & 86.32 &17.74 & 3.64 
    & 97.02 & 22.68 & 52.73 & 86.37 & 18.0 & 12.18 \\
\bottomrule
\end{tabular}
}
\end{table*}
\paragraph{Optimal bias vs. model architectures.}
To further validate that different tasks require different degrees of simplicity bias, we examine how the optimal $\beta$ interacts with the inherent bias of three representative models. Our results show that the optimal choice of $\beta$ is also influenced by the simplicity bias induced by model architectures.
 As observed in Section~\ref{sec:effect}, ViT exhibits a stronger simplicity bias than ResNets, which is also reflected in the direction of bias modulation: for instance, while ResNets benefit from enhancing simplicity bias on the iWildCam dataset, ViT-B/16 requires suppressing it. This supports our hypothesis that {a ``simple'' model needs to increase its complexity to better match the requirements of a ``complex'' task}.  A similar trend is also observed under adversarial attack evaluations (Section~\ref{sec:attack}).

When comparing models from the same architecture family, such as RN50 and RN101, we observe that their optimal $\beta$ values are similar, even though RN50 exhibits a stronger simplicity bias (Table~\ref{tab:complexity-zeroshot}). Based on the insights in Section~\ref{sensitivity-performance}, we speculate that the benefit of tuning $\beta$ for RN101 may lie in its increased sensitivity to low-frequency components, which compensates for its weaker inherent simplicity bias relative to RN50. In adversarial settings (Section~\ref{sec:attack}), where the frequency distribution of inputs is less constrained, the difference between RN50 and RN101 becomes more apparent and aligns with their inherent biases.

\begin{table}[t]
\centering
\caption{
Accuracy (\%) of $\gamma_s$-parameterized ViT-B/16 under various adversarial attacks. Increasing $\gamma_s$ improves robustness under gradient-based attacks, while robustness to score-based attacks peaks at lower $\gamma_s$.
}
\label{tab:adv_vit}

\renewcommand{\arraystretch}{1.2}
\resizebox{\linewidth}{!}{%
\begin{tabular}{c c cccc c}
\toprule
\multirow{2}{*}{$\gamma_s$} & \multirow{2}{*}{Clean} 
& \multicolumn{4}{c}{\textbf{Gradient-based}} 
& \textbf{Score-based} \\
\cmidrule(lr){3-6} \cmidrule(lr){7-7}
& & PGD & FGSM & BIM& MIFGSM & Pixle \\
\midrule
1.5 & 97.74 & \textbf{32.54} &\textbf{ 50.81} & \textbf{87.54} & \textbf{20.94} & 90.57 \\
1.2 & 97.85 & 22.62 & 44.31 & 85.59 & 15.31 & 91.52 \\
1.0 & \textbf{98.12} & 21.72 & 49.43 & 83.17 & 16.71 & \textbf{92.59} \\
0.9 & 98.04 & 17.33 & 46.24 & 80.40 & 13.47 & 90.61 \\
\bottomrule
\end{tabular}

}
\end{table}

\begin{table}[t]
\centering
\caption{
Accuracy (\%) under transfer attacks on RN101. The adversarial examples are crafted using RN50  with varying $\beta$ via PGD and FGSM attacks, both under an $\ell_\infty$ perturbation of 0.03. $\Delta$ Acc denotes accuracy drop compared to $\beta=1$.
}
\label{tab:trans_attack50}
\begin{tabular}{lcc|c}
\toprule
\multirow{2}{*}{Surrogate Model} & \multicolumn{2}{c|}{Attack Method} & \multirow{2}{*}{$\Delta$ Acc} \\
                                 & PGD            & FGSM           &                             \\ \midrule
RN50 ($\beta=1.0$)              & 55.30                & 61.87                & 0.00                        \\
RN50 ($\beta=0.9$)              & 54.80                & 61.11                & 0.63                        \\
RN50 ($\beta=0.8$)              & 54.36                & 61.24                & 0.79                        \\
RN50 ($\beta=0.7$)              & \textbf{53.54}                & \textbf{60.61}                & \textbf{1.51}                        \\
\bottomrule
\end{tabular}
\end{table}

\begin{table}[t]
\centering
\caption{Accuracy (\%) under transfer attacks on RN50. Adversarial examples are generated using RN101 with varying $\beta$ via PGD and FGSM attacks, both under an $\ell_\infty$ perturbation of 0.03.
 $\Delta$ Acc denotes accuracy drop  compared to $\beta=1$.}
\label{tab:trans_attack101}
\begin{tabular}{lcc|c}
\toprule
\multirow{2}{*}{Surrogate Model} & \multicolumn{2}{c|}{Attack Method } & \multirow{2}{*}{$\Delta$ Acc} \\
                                 & PGD            & FGSM           &                             \\ \midrule
RN101 ($\beta=1.0$)              & 53.85                & 55.87                & 0.00                        \\
RN101 ($\beta=0.9$)              & 54.43                & 56.79                & -0.75                       \\
RN101 ($\beta=0.8$)              & {52.95}       & 55.69                & 0.54                        \\
RN101 ($\beta=0.7$)              & \textbf{52.75}                & \textbf{54.99}       & \textbf{0.99}                        \\
\bottomrule
\end{tabular}
\end{table}
\section{Simplicity Bias in Adversarial Robustness}
\label{sec:attack}

We investigate the effect of simplicity bias on model robustness against various adversarial attacks. We use the same ResNet-50, ResNet-101, and ViT-B/16 weights fine-tuned on CIFAR-10, as described in Section~\ref{basicsetting}. Results for ResNets and ViT-B/16 are reported in Table~\ref{tab:adv_rn} and Table~\ref{tab:adv_vit}, respectively.
\subsection{Attack Settings}
We consider two types of adversarial attacks: (1) gradient-based attacks, including PGD~\cite{madry2017towards}, FGSM~\cite{goodfellow2014explaining}, BIM~\cite{kurakin2018adversarial}, and MIFGSM~\cite{dong2018boosting}; and (2) a score-based attack, Pixle~\cite{pomponi2022pixle}.
For all gradient-based attacks,  the perturbation magnitude is set to $\epsilon = 0.03$ under the $\ell_\infty$ normnorm, while Pixle is configured with both the $x$- and $y$-dimensions set to $(0.1, 0.4)$, a single restart, and 50 iterations; all other parameters follow their default settings. All attacks are implemented using the Torchattacks library~\cite{kim2020torchattacks}.
Results are averaged over 5000 adversarial examples. 
\begin{table*}[t!]
\centering
\small
\caption{
Accuracy of RN50 under corruptions from CIFAR-10-C benchmark. Numbers are filled from raw results. The best average accuracy is achieved at  $\beta=0.8$.
}
\label{tab:cifarc_rn50}
\vspace{0.3em}
\begin{tabular}{c|c|c c c c c c c c c|c}
\toprule
Model & $\beta$ & Snow & Brightness & \makecell{Impulse\\Noise} & Frost & Contrast & Fog & \makecell{Shot\\Noise} & \makecell{Defocus\\Blur} & \makecell{JPEG\\compression} & Avg \\
\midrule
\multirow{4}{*}{RN50}
& 1.0 & 88.40 & 94.79 & 55.85 & 84.24 & 88.90 & 93.51 & 38.64 & 92.31 & \textbf{80.63} & 79.70 \\
& 0.9 & 88.16 & 95.30 & 54.49 & 83.10 & 91.19 & 94.11 & 35.18 & \textbf{93.75} & 79.81 & 79.45 \\
& 0.8 & \textbf{89.27} & \textbf{95.69} & \textbf{63.33} & \textbf{83.92} & \textbf{91.69} & \textbf{94.44} & \textbf{38.80} & 93.70 & 80.43 & \textbf{81.25} \\
& 0.7 & 88.55 & 95.65 & 55.39 & 82.59 & 91.37 & 94.43 & 26.22 & 93.93 & 79.56 & 78.63 \\
\bottomrule
\end{tabular}
\end{table*}

\begin{table*}[t!]
\centering
\small
\caption{
Accuracy of RN101 under corruptions from CIFAR-10-C benchmark. The best average accuracy is achieved at $\beta=0.8$.
}
\label{tab:cifarc_rn101}
\vspace{0.3em}
\begin{tabular}{c|c|c c c c c c c c c|c}
\toprule
Model & $\beta$ 
& Snow & Brightness & \makecell{Impulse\\Noise} & Frost & Contrast & Fog & \makecell{Shot\\Noise} & \makecell{Defocus\\Blur} &  \makecell{JPEG\\compression} & Avg \\
\midrule
\multirow{4}{*}{RN101}
& 1.0 & 91.46 & 95.60 & 65.34 & 86.67 & 91.72 & 94.58 & 55.53 & 93.89 & \textbf{83.03} & 84.20 \\
& 0.9 & 91.15 & 95.40 & 72.40 & 86.23 & \textbf{92.83} & \textbf{95.22} & \textbf{57.97} & 93.44 & 80.03 & 84.96 \\
& 0.8 & 91.21 & \textbf{96.23} & \textbf{72.73} & \textbf{88.45} & 92.46 & 94.72 & 57.81 & \textbf{95.38} & 81.85 & \textbf{85.65} \\
& 0.7 & \textbf{92.15} & 95.80 & 68.15 & 87.13 & 92.53 & 94.31 & 47.44 & 94.98 & 80.97 & 83.72 \\
\bottomrule
\end{tabular}
\end{table*}

\begin{table*}[t!]
\centering
\small
\caption{
Accuracy of ViT-B/16 under corruptions from CIFAR-10-C benchmark. Best accuracy is achieved at $\gamma=1.0$.
}
\label{tab:imagenet_c_vit_gamma}
\vspace{0.3em}
\begin{tabular}{c|c|c c c c c c c c c|c}
\toprule
Model & $\gamma$ & Snow & Brightness & \makecell{Impulse\\Noise} & Frost & Contrast & Fog & \makecell{Shot\\Noise} & \makecell{Defocus\\Blur} &  \makecell{JPEG\\compression} & Avg \\
\midrule
\multirow{4}{*}{ViT-B/16}
& 1.5 & 94.27 & 97.14 & 88.08 & 92.35 & 95.16 & 96.04 & 68.16 & 95.98 & 86.87 & 90.34 \\
& 1.2 & 93.98 & 97.41 & 88.59 & 92.28 & 95.56 & 96.56 & 69.25 & 96.31 & \textbf{87.70} & 90.85 \\
& 1.0 & \textbf{94.89} & \textbf{97.70} & 87.44 & \textbf{93.87} & \textbf{96.06} & \textbf{96.67} & \textbf{72.44} & \textbf{96.75} & 87.30 & \textbf{91.35} \\
& 0.9 & 94.57 & 97.56 & \textbf{89.13} & 93.15 & 95.89 & 96.61 & 63.11 & 96.78 & 87.31 & 90.46 \\
\bottomrule
\end{tabular}
\end{table*}

\subsection{Main Results}


From Table~\ref{tab:adv_rn}, we observe that, unlike OOD generalization, robustness under most adversarial attacks tends to improve with higher model complexity. Table~\ref{tab:adv_vit} further supports this observation: increasing $\gamma_s$ to 1.5 consistently enhances robustness against PGD, FGSM, BIM, and MIFGSM attacks by a clear margin.
As noted by Madry et al.~\cite{madry2017towards}, classifying adversarial examples requires more complex decision boundaries than natural images, which aligns with our findings.

However, robustness against score-based attacks (Pixle) shows lower $\beta$ or $\gamma_s$ can still improve robustness, implying that different attack types interact with simplicity bias differently. 

Furthermore, we examine the preferences across models. Since simplicity bias cannot be reduced in ResNets, a fair comparison with ViT on gradient-based attacks is not possible. However, we can still compare their robustness under the Pixle attack.
From this, we conclude that ViT-B/16 requires less simplicity bias than any ResNet.  
Within ResNets, RN101 requires the strongest modulation of simplicity bias.  
This trend aligns with their simplicity bias after fine-tuning, as shown in Table~\ref{tab:freq_sens}.



\section{ Simplicity Bias in Robustness to Transfer Attacks}
The transferability of adversarial examples is known to depend on the similarity in feature representations and inductive biases between the surrogate and target models~\cite{ilyas2019adversarial,ma2024improving}. To examine whether simplicity bias influences adversarial transfer, we investigate how the frequency preferences of surrogate models affect attack effectiveness on the CIFAR-10 dataset.
\subsection{Evaluation Settings}
We use the same fine-tuned CLIP models with ResNet-50 and ResNet-101 backbones as in Section~\ref{basicsetting}. For transfer attack evaluation, one model acts as the surrogate model to generate adversarial examples on CIFAR-10 images, while the other serves as the target model. Surrogate models are varied in simplicity bias by adjusting $\beta$, whereas the target model remains at its original setting ($\beta=1$). We generate 5,000 adversarial examples for each surrogate model to assess transferability. Two attack methods are considered: PGD and FGSM, both with $\ell_\infty$ bounds of 0.03. Results are reported in terms of classification accuracy and accuracy drop ($\Delta$Acc) relative to the $\beta = 1$ baseline. Tables~\ref{tab:trans_attack50} and~\ref{tab:trans_attack101} report the results for RN50 and RN101 as target models, respectively.

\paragraph{Main results.} In both experiments, surrogate models with the strongest simplicity bias ($\beta=0.7$) yield the most effective transfer attacks. e attribute this to models with stronger simplicity bias inducing perturbations that transfer more effectively across models.  An exception at $\beta = 0.9$ in Table~\ref{tab:trans_attack101} suggests that increased robustness in the surrogate model can hurt transferability.

\section{ Simplicity Bias in Image Corruption Robustness}
We analyze the impact of simplicity bias on robustness to image corruptions using the CIFAR-10-C benchmark~\cite{hendrycks2019benchmarking}. Our evaluation covers nine corruption types at severity level 3, including additive noise, blur, contrast, and compression.  We use the same fine-tuned CLIP models with ResNet-50, ResNet-101, and ViT-B/16 backbones as in Section~\ref{basicsetting}. Accuracy with varying simplicity bias on the full CIFAR-10-C test set (10,000 images) is reported in Tables~\ref{tab:cifarc_rn50}, \ref{tab:cifarc_rn101}, and \ref{tab:imagenet_c_vit_gamma}, respectively.
\paragraph{Main results.}
We observe that $\beta = 0.8$ yields the highest average accuracy across the nine corruptions for both RN50 and RN101, whereas ViT-B/16 performs best at $\beta = 1.0$. Compared to the optimal values on the OOD benchmark ($0.7$, $0.7$, and $0.9$), robustness to image corruptions generally benefits less from additional simplicity bias. One explanation is that CIFAR-10-C corruptions span a broader frequency spectrum than clean images~\cite{yin2019fourier}, making excessive SB less effective. 




\section{Conclusions}
\noindent\textbf{Contributions.} In this work, we investigate the role of simplicity bias in fine-grained image classification tasks, including zero-shot, OOD, adversarial and transfer attacks, and image corruptions. We provide a theoretical analysis that identifies gaps in prior studies and propose a frequency-aware metric capable of capturing simplicity bias changes induced by activation smoothness and LayerNorm scaling. This metric also reveals that the model tends to learn low-frequency features when SB is strengthened. Our empirical results suggest that task performance tends to be higher when the simplicity bias aligns with task-specific preferences. These findings advance the understanding of simplicity bias and may guide future improvements in generalization.

\noindent\textbf{Limitations.} While we focus on simplicity bias and control for other variant factors like model capacity, training influences, etc., as noted by Teney et al.~\cite{Teney_2025_CVPR}, simplicity bias is only one aspect of inductive bias; thus, two models with different simplicity biases can still have the same performance. Identifying other influential biases would further contribute to the understanding of generalization and robustness. 


\noindent\textbf{Future work.} Our simplicity bias metric can serve as a supervisory signal during training, allowing learnable adjustment of the bias to better suit downstream tasks. This improves performance without requiring architectural modifications or additional data augmentations.

\bibliographystyle{IEEEtran}
\bibliography{egbib}
\end{document}